\documentclass[letterpaper, 10 pt, conference]{ieeeconf}  

\usepackage{times}
\usepackage{amsmath,amsfonts}
\usepackage{mathtools}
\usepackage{float}
\usepackage{textcomp}
\usepackage{multicol}
\usepackage{multirow}
\usepackage[bookmarks=true]{hyperref}
\usepackage{balance}
\usepackage[short,c2]{optidef}

\newtheorem{lemma}{Lemma}
\newtheorem{corollary}{Corollary}
\newtheorem{theorem}{Theorem}
\usepackage{makeidx}  
\makeindex
\usepackage[noadjust]{cite}
\usepackage{url}

\usepackage{xcolor}
\usepackage{colortbl}
\usepackage{placeins}
\usepackage{booktabs}
\usepackage{siunitx}
\IEEEoverridecommandlockouts                              

\title{\LARGE \bf
Ergodic Exploration over Meshable Surfaces
}

\author{Dayi Dong$^{1}$, Albert Xu$^{2}$, Geordan Gutow$^{2}$, Howie Choset$^{2}$, Ian Abraham$^{3}$
\thanks{$^{1}$Dayi Dong is with the Department of Mechanical Engineering, University of California Berkeley, Berkeley, CA 94709, USA {\tt \small dayi.dong@berkeley.edu}}%
\thanks{$^{2}$Albert Xu, Geordan Gutow, and Howie Choset are with the Robotics Institute, Carnegie Mellon University, 5000 Forbes Ave., Pittsburgh, PA 15213 {\tt \small \{albertx, ggutow, choset\}@andrew.cmu.edu}}%
\thanks{$^{3}$Ian Abraham is with the Department of Mechanical Engineering and Materials Science, Yale University, New Haven 06511, USA. {\tt \small ian.abraham@yale.edu}}%
\thanks{*This research was supported in part by an Intelligence Community Postdoctoral Research Fellowship at Carnegie Mellon University, administered by Oak Ridge Institute for Science and Education through an interagency agreement between the U.S. Department of Energy and the Office of the Director of National Intelligence.}
}

\begin{document}

\maketitle

\begin{abstract}

Robotic search and rescue, exploration, and inspection require trajectory planning across a variety of domains. A popular approach to trajectory planning for these types of missions is ergodic search, which biases a trajectory to spend time in parts of the exploration domain that are believed to contain more information. Most prior work on ergodic search has been limited to searching simple surfaces, like a 2D Euclidean plane or a sphere, as they rely on projecting functions defined on the exploration domain onto analytically obtained Fourier basis functions.  In this paper, we extend ergodic search to any surface that can be approximated by a triangle mesh. The basis functions are approximated through finite element methods on a triangle mesh of the domain. We formally prove that this approximation converges to the continuous case as the mesh approximation converges to the true domain. We demonstrate that on domains where analytical basis functions are available (plane, sphere), the proposed method obtains equivalent results, and while on other domains (torus, bunny, wind turbine), the approach is versatile enough to still search effectively.  Lastly, we also compare with an existing ergodic search technique that can handle complex domains and show that our method results in a higher quality exploration.

\end{abstract}

\section{INTRODUCTION}

\begin{figure}[h!]
    \centering
    \includegraphics[width=.8\linewidth]{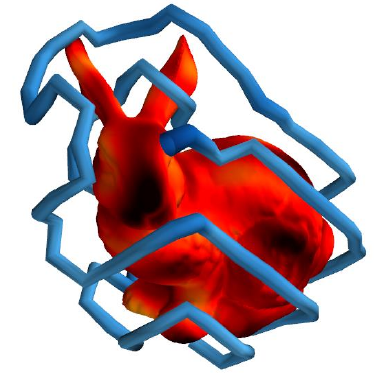}
    \caption{\textbf{Ergodic Search over a Complex Meshable Surface} The proposed approach plans dynamically feasible trajectories that search with respect to an information map over any surface that a triangle mesh can approximate. Shown is an inspection trajectory for the Stanford bunny with a uniform information map.
    }
    \label{fig:1}
    \vspace{-20pt}
\end{figure}

Planning a robot trajectory to gather information is central to tasks such as terrain exploration, structural inspection, and search and rescue \cite{waharte2010supporting, mayer2019drones, ruangpayoongsak2005mobile}.  One option is to simply visit the entire domain, but when \textit{a priori} knowledge about the distribution of information in the exploration domain exists, a planner can leverage this information to improve search performance. Doing so requires sophisticated planning techniques \cite{jacobs2010multiscale,silverman2013optimal,mathew2011metrics}.

One particularly popular approach to \textit{information-driven} search is ergodic search \cite{mathew2011metrics,miller2013trajectory,miller2015ergodic,coffin2022multi,lerch2023safety,dong2023time}. Intuitively, in ergodic search, one generates a trajectory such that time spent in a region is proportional to the amount of information in that area according to the prior knowledge (referred to as an \textit{information map}). Ergodic search allows us to plan in continuous space and effectively balances the search for new information (exploration) and the utilization of existing information (exploitation). 
    
Classic ergodic search relies on comparing the Fourier transforms of a trajectory and the information map, thus, the space being explored (the \textit{exploration domain}) has typically been limited to simple surfaces with available analytical Fourier basis functions \cite{mathew2011metrics,miller2013trajectory, jacobs2010multiscale}. One recent approach can generate ergodic trajectories without the use of analytical Fourier basis functions by solving a partial differential equation (PDE) on the volume surrounding the domain of interest \cite{ivic2023multi}. This Heat Equation-Driven Area Coverage (HEDAC) method is very general and versatile, extending to multiple robots, but at the expense of needing to solve a PDE. This PDE depends on the information map and so must be resolved if the information map changes (even if the domain has not). We also observe that short trajectories do not achieve good coverage.

Due to these limitations, we seek to instead generalize the classic ergodic search approach to domains where analytical basis functions are not available.
The core contribution of this paper is a method of performing ergodic trajectory optimization (ETO) over any 2D manifold that can be approximated by a triangle mesh. We do so by computing Laplace-Beltrami eigenfunctions (a generalization of the Fourier basis functions) \cite{Sharp:2020:LNT} for the mesh approximation of the manifold and conducting conventional ergodic trajectory optimization \cite{lerch2023safety} using these approximate basis functions to generate trajectories like in Fig. \ref{fig:1}.

    Section \ref{sec:related} describes related works in uniform coverage methods and ergodic search.
    Section \ref{sec:erg_search} covers mathematical preliminaries and describes the classic ergodic search approach.
    Section \ref{sec:search_surfaces} formalizes the mesh approximation we use here, including a proof that the approximate objective converges to the original objective as the mesh is refined, and explains how the mesh approximations can be used to formulate a trajectory optimization problem.
    Section \ref{sec:results} gives the implementation details for our experiments, in which we compare to using analytical basis functions on simple surfaces and to HEDAC.
    Section \ref{sec:conclusion} outlines the conclusions we draw and potential future work.

\section{Related Work} \label{sec:related}

    \subsection{Uniform Coverage-Based Methods}
    Uniform coverage-based methods seek to examine the largest possible fraction of a bounded domain \cite{galceran2013survey, choset2001coverage} and are appropriate for exploration and mapping when the goal is to visit the entire exploration space \cite{maxim2012uniform}. Coverage methods often discretize the entire domain and plan a trajectory that visits each cell \cite{yap2002grid}. The shortest such trajectory can be found by solving a traveling salesman problem \cite{flood1956traveling}, for which many algorithms have been proposed \cite{applegate1998solution}. It is also common to use exploration-specific heuristics like the Lawnmower algorithm \cite{ousingsawat2007modified}. Other coverage-based methods have instead searched over continuous domains by utilizing spatial approximations like limited Voronoi partitioning \cite{pratissoli2022coverage}, sensors that cover a certain limited region \cite{bouman2022adaptive}, or attractive and repulsive potential fields \cite{howard2002mobile, kim2006local}. 
    
    Uniform coverage approaches specific to 3D have been proposed, often with a focus on real-world inspection tasks. One early work established how to generate a set of viewpoints that collectively cover all portions of a surface but did not fully address how a robot would move to the viewpoints \cite{pito1999solution}.  Other methods simply approximated the exploration domain as a simpler space (eg. 2D Euclidean space) \cite{cho2021coverage,de2005blind} when adding additional nuance would not benefit the problem definition or solution generation. 
    
    The idea of searching through prepositioned waypoints has been further explored \cite{hover2012advanced, bircher2016three, janouvsek2013speeding}. Additional work has explored re-meshing techniques \cite{alexis2015uniform} and more realistic sensors \cite{dornhege2013coverage}. 
    Non-waypoint methods have been proposed for 3D, including planar terrain-covering algorithms \cite{hert1996terrain}, 3D cellular decomposition \cite{atkar2001exact}, and random sampling algorithms \cite{englot2012sampling}.

    \subsection{Ergodic Search}
    Uniform coverage approaches generally do not exploit \textit{a priori} knowledge of the distribution of \textit{information} in the exploration domain to bias the search towards areas of greater interest. This information could be the likelihood of a survivor or failure at a location, so it is desirable to exploit prior knowledge if available. Information-driven search, of which ergodic search is a popular example, is one way to do so. 
    Ergodic search is appropriate when prior knowledge about the non-uniform distribution of information is available \cite{choset2001coverage, mathew2011metrics}, as it biases the robot to spend longer in areas of high information and less in areas of low information. 
    
    The key tool in ergodic search is the ergodic metric \cite{mathew2011metrics}, a scalar that quantifies the efficiency of a trajectory at exploring the area with respect to the prior knowledge (encoded as a probability density function over the exploration domain, referred to as an \textit{information map}). This approach has been shown to be highly beneficial for information-gathering tasks \cite{silverman2013optimal, dressel2018optimality} and has been proposed for numerous applications. Previous work has incorporated collision avoidance \cite{lerch2023safety}, considered platforms with complex dynamics \cite{dong2023time}, and studied various sensor models (eg. low-information sensors \cite{coffin2022multi}). One limitation of the existing ergodic search literature is that it has usually only been conducted over simple surfaces with analytical basis functions \cite{jacobs2010multiscale, miller2013trajectory,alexis2015uniform}. 

    Recent work has utilized the HEDAC algorithm to generate ergodic trajectories in 3D for inspection of surfaces without requiring analytical basis functions \cite{ivic2023multi}. The algorithm was originally presented for 2D in \cite{ivic2016ergodicity}. This approach can provide ergodic trajectories for a variety of surfaces but must solve a PDE for potential fields that yield the ergodic trajectories. This is extremely computationally expensive and, for short trajectories, achieves worse coverage than ergodic trajectory optimization approaches like \cite{lerch2023safety}. In this work, we address these limits by directly optimizing for trajectories that minimize the ergodic metric over a finite time horizon using direct collocation \cite{lerch2023safety}.

\section{Preliminaries on Ergodic Search} \label{sec:erg_search}


    With $\mathcal X$ being the robot state space, let $x:\mathbb R\to\mathcal X$ or $x(t)$ denote a trajectory in time through $\mathcal X$. Let $\mathcal{U}$ be the set of controls, and $u(t)$ denote a trajectory in time through $\mathcal{U}$. Next, let $\mathcal W$ be the exploration domain that we seek to search over. In most prior work, this is a bounded $n$-dimensional hypercube $\mathcal W=[0,L_1] \times [0,L_2]...\times [0,L_n]\subset \mathbb R^n$, but we extend this to any manifold embedded in $R^{n}$ that may be approximated by a homogeneous simplicial complex (a triangular mesh in $\mathbb{R}^3$). Our application is the 3D case, and the implementation we present is limited to 3D, so we will often abuse terminology and speak specifically of a triangular mesh rather than the more general homogeneous simplicial complex.
    By approximating a manifold as a triangular mesh, we mean that there exists a sequence of triangular meshes for which the Hausdorff distance between mesh and manifold converges to 0 (see \cite{hildebrandt2006convergence} for details).
    
The space $\mathcal X$ and $\mathcal W$ are connected by a sensor model $s^{x}(w):\mathcal X\times\mathcal W\to \mathbb R$ that collects information density $s^x(w)$ at the point $w\in\mathcal{W}$ when the robot is at state $x\in\mathcal{X}$.
    The sensor model must be positive $s^x(w)>0$, but the total information gathered at a point $x$ need not be normalized. It describes how "well" the robot can "see" each point in $\mathcal W$ when at a particular state.
    In classic ergodic search, $s^x$ is a Dirac delta function $\delta_x$; the robot can see only its current location.

    Rolling out the robot's trajectory $x(\cdot)$, we define the trajectory's unnormalized time-averaged statistics $\hat{\mu}^{x(\cdot)}(w)$ in Eq.~\eqref{eq:time-average-statistics-unnormalized} as how much information is gathered at each point in $\mathcal W$:

    \begin{equation}
        \hat{\mu}^{x(\cdot)}(w) = \frac{1}{t_f} \int_{0}^{t_f} s^{x(t)}(w) dt \label{eq:time-average-statistics-unnormalized}\\
    \end{equation}
The unnormalized statistics can be treated as a distribution $\mu^{x(\cdot)}(w)$ by normalizing with respect to $\mathcal W$ in Eq.~\eqref{eq:time-average-statistics}.
\begin{equation}
    \mu^{x(\cdot)}(w) = \frac{1}{\int_\mathcal{W} \hat{\mu}^{x(\cdot)}(w)dw} \hat{\mu}^{x(\cdot)}(w)  \label{eq:time-average-statistics}
\end{equation}

    Next, in the infinite time limit $t_f\to\infty$, a trajectory $x(t)$ can be called ergodic over $\mathcal W$ if the information it gathers, $\mu^{x(\cdot)}(w)$, converges weakly to the desired measure of information $\phi: \mathcal{W} \rightarrow \mathbb{R}^+$. $\phi$ will often be referred to as the \textit{information map}. Such convergence indicates that the trajectory spends search effort on a region proportional to the information in that region.
    Formally, a trajectory $x(t)$ is ergodic if Eq.~\eqref{eq:erg_def} holds for all Lebesgue integrable functions $f \in \mathcal{L}^1$ \cite{scott2009capturing}. 
    Note that the left-side expression is the inner product $\langle \lim_{t_f\to\infty}\mu^{x(\cdot)}(w),\mu(w)\rangle$.
    \begin{equation} \label{eq:erg_def}
        \lim_{t_f\to\infty} \int_{\mathcal W}f(w) \mu^{x(\cdot)}(w)dw = \int_\mathcal{W} f(w) \phi(w) dw
    \end{equation}

    In finite time, we do not expect to drive $\mu^x$ exactly to $\phi$. To quantify the difference between the desired measure and the trajectory's time-averaged statistics, we take a weighted squared norm between $\phi(w)$ and $\mu^{x(\cdot)}(w)$ under a spectral basis.
    In Eq.~\eqref{eq:ergodic_met}, $\mu_k$ and $\phi_k$ are the $k$th basis coefficients of the $\mu^{(\cdot)}$ and $\phi$ distributions respectively, computed using the basis functions $f_k$. $\Lambda_k$ is a weighting term that discounts higher-frequency modes. In the special case where $\mathcal W$ is a hypercube, the basis functions match the Fourier basis functions in Eq.~\eqref{eq:fourier-bases}, and thus, $\mu_k$ and $\phi_k$ are the $k$th Fourier coefficients.
    
    \begin{gather}\label{eq:ergodic_met}
        \mathcal{E}(x(t), \phi) = \sum_{k=0}^K \Lambda_k \left( \mu_k - \phi_k \right)^2 \\
        \mu_k = \int_{\mathcal W} f_k(w) \mu^{x(\cdot)}(w)dw \label{eq:ck} \\
        \phi_k = \int_{\mathcal W} f_k(w) \phi(w)dw \label{eq:phik}
    \end{gather}

    \begin{equation} \label{eq:fourier-bases}
        f_k(w) = \frac{1}{h_k} \prod_{i=1}^{n} \cos\left(\frac{w_i k_i \pi}{L_i}\right)
    \end{equation}

    In ergodic search \cite{mathew2011metrics}, one seeks a robot trajectory that minimizes $\mathcal{E}$. The primary contribution of this paper lies in the choice of basis functions $f_k$ for computing $\mathcal{E}$ on manifolds $\mathcal W$ that can be approximated as a triangle mesh.

\section{Search on Surfaces} \label{sec:search_surfaces}    
    \subsection{Approximation of Basis Functions} \label{subsec:basis_approx}
    The key theoretical difficulty for extending the ergodic metric to manifolds is selecting basis functions on $\mathcal{W}$. We require that they form an orthonormal basis ($\int_\mathcal{W} f_i f_j dw = \delta_{ij}$) to ensure efficient calculation of the $\mu_k$ and $\phi_k$ coefficients.
    In addition, we discount the higher-order basis functions using $\Lambda_k$ to ensure the sum in Eq.~\eqref{eq:ergodic_met} converges even as $K\to\infty$, so we prefer the basis functions to have some relation to frequency.

    For Euclidean spaces, the analytical Fourier basis functions Eq.~\eqref{eq:fourier-bases} satisfy our conditions, but for an arbitrary manifold $\mathcal W$, they fail orthonormality.
    As the Fourier bases are the eigenfunctions of the Laplacian operator, we thus turn to the Laplace-Beltrami operator ($\Delta$), which generalizes the Laplacian to $\mathcal W$.
    Some prior work has explored using the Laplace-Beltrami eigenfunctions for ergodic search \cite{xu2024measure}, but we formalize their use in this paper. Because $\Delta$ is Hermitian, its eigenfunctions Eq.~\eqref{eq:eigenvalue-continuous} will be orthonormal, and the respective eigenvalues will be related to the frequency of the eigenfunction \cite{Sharp:2020:LNT}. 
    \begin{equation}\label{eq:eigenvalue-continuous}
        \Delta f_k = \lambda_k f_k
    \end{equation}
    In Euclidean space, the eigenfunctions of $\Delta$ coincide with the Fourier basis functions, making our proposed method consistent with prior approaches.
    In practice, we need to approximate $\Delta$ using a discretization. In this work, we use the Laplace-Beltrami operator $L$ of a simplicial manifold $\mathbb{M}$ that approximates the original manifold $\mathcal{W}$. Specifically, in $\mathbb{R}^3$, $\mathbb{M}$ is a triangular mesh, and we restrict $\mathcal{W}$ to be a manifold for which there exists a sequence of triangle meshes which 1.) converge to $\mathcal{W}$ in Hausdorff distance 2.) converge to $\mathcal{W}$ in area. and 3.) are \textit{normal graphs} over $\mathcal{W}$ (a technical condition, see \cite{hildebrandt2006convergence}). This restriction on $\mathcal{W}$ ensures that $L\to{}\Delta$ in the operator norm \cite{hildebrandt2006convergence}.
    
    For a triangle mesh with a finite number of vertices $m$, $L$ is an $m\times m$ matrix that can be computed using eg. \cite{Sharp:2020:LNT}. Functions $f:\mathcal{W}\to \mathbb{R}$ are approximated on $\mathbb{M}$ as length $m$ vectors by integrating over the area element $E_i\subset{}\mathbb{M}$ associated with vertex $v_i$: $\vec f = [\int_{E_i}f(v)dv\approx{}f(v_i)\text{Area}(E_i)]_{i=1}^m$. This gives us the following matrix eigenvalue problem analogous to Eq.~\eqref{eq:eigenvalue-continuous}.
    \begin{equation} \label{eq:eigenvalue-approx}
        L \vec{f_k} = \lambda_k\vec{f_k}
    \end{equation}
Next, we show that as the mesh $\mathbb{M}$ converges to $\mathcal{W}$, so too does the ergodic metric computed on $\mathbb{M}$ using the eigenvectors from ~\eqref{eq:eigenvalue-approx} converge to the metric we would have on $\mathcal{W}$ using the eigenfunctions from Eq.~\eqref{eq:eigenvalue-continuous}.
    \begin{lemma} \label{lemma:normal-op}
        If $A$ is a Normal operator and $g$ a function, the inner product $\langle g, Ag\rangle$ equals $\sum_{k} \lambda_k g_k^2$, where $\lambda_k$, $f_k(x)$ are the $k$th eigenvalue/eigenfunction of $A$ and $g$ is decomposed as $g(x) = \sum_k g_k f_k(x)$.
    \end{lemma}
    \begin{proof}
        By the spectral theorem, $A$ admits an orthonormal basis $f_k(x)$ with (real) eigenvalues $\lambda_k$. This allows us to write $g$ as a sum over $f_k$ with coefficients $g_k = \langle g,f_k\rangle$.
        \begin{equation}
            g(x) = \sum_k g_k f_k(x)
        \end{equation}

        Since $A$ is linear and the bases $f_k(x)$ are orthonormal, we obtain the result with some algebra.
        \begin{align}
            \langle g, Ag\rangle &= \sum_{j,k} g_j g_k \langle f_j, A f_k\rangle \\
            &= \sum_{j,k} g_j g_k \lambda_k \langle f_j, f_k\rangle \\
            &= \sum_k \lambda_k g_k^2
        \end{align}
    \end{proof}

    \begin{corollary} \label{coro:lamk-power-series}
        If $\eta(\lambda_k)$ is a power series in $\lambda_k$, then $\sum_k \eta(\lambda_k) g_k^2 = \langle g, \eta(A) g\rangle$.
    \end{corollary}
    \begin{proof}
        First, write $\eta(\lambda_k)$ as a power series
        \begin{equation}
            \eta(\lambda_k) = \sum_{n=0}^\infty a_n \lambda_k^n
        \end{equation}

        Using Lemma \ref{lemma:normal-op}, we have
        \begin{align}
            \sum_k  \sum_{n=0}^\infty a_n \lambda_k^n g_k^2 &= \sum_{n=0}^\infty a_n \left\langle g, A^n g\right\rangle \\
            &= \left\langle g, \sum_{n=0}^\infty a_n A^n g\right\rangle \\
            &= \langle g, \eta(A) g\rangle
        \end{align}

        where $\eta(A)$ is defined by its power series.
    \end{proof}

    \begin{corollary} \label{coro:ergodicity-inner-product}
        Let the ergodic discounting factor $\Lambda_k=\eta(\lambda_k)$ be a power series. Then the ergodic metric calculated using the continuous Laplace-Beltrami operator $\Delta$ is
        \begin{equation}\label{eq:erg_met_cont}
            \mathcal E^\Delta = \left\langle \phi-\mu^{x(\cdot)}, \eta(\Delta)(\phi-\mu^{x(\cdot)})\right\rangle
        \end{equation}

        and the ergodic metric calculated using the discretized operator $L$ is
        \begin{equation}\label{eq:erg_met_disc}
            \mathcal E^L = \left\langle \phi-\mu^{x(\cdot)}, \eta(L)(\phi-\mu^{x(\cdot)})\right\rangle
        \end{equation}
    \end{corollary}
    \begin{proof}
        Because $\Delta$ and $L$ are both Normal, the proof follows from Corollary \ref{coro:lamk-power-series} with $g=\phi-\mu^{x(\cdot)}$.
    \end{proof}

    \begin{theorem}
    The ergodic metric computed using the discrete operator $L$ converges to the continuous metric computed using $\Delta$ as the mesh discretizations $\mathbb{M}_n$ approach $\mathcal W$.
    \end{theorem}
    \begin{proof}
        From Corollary \ref{coro:ergodicity-inner-product}, we have the error between ergodic metrics $\mathcal E^\Delta$ and $\mathcal E^L$.
        \begin{align}
            \left|\mathcal E^\Delta - \mathcal E^L\right| &= \left|\left\langle \phi-\mu^{x(\cdot)}, (\eta(\Delta) - \eta(L))(\phi-\mu^{x(\cdot)})\right\rangle\right| \\
            &\leq \|\eta(\Delta) - \eta(L)\|_\text{op} \|\phi-\mu^{x(\cdot)}\|^2
        \end{align}

        Because $\eta$ is continuous, $L$ converging to $\Delta$ implies that $\eta(L)\to\eta(\Delta)$ in the operator norm. Therefore, by the above inequality, $\mathcal E^L$ converges to $\mathcal E^\Delta$ as $L\to\Delta$. From the results of Hildebrandt et al. \cite{hildebrandt2006convergence}, $L\to\Delta$ as a sequence of meshes converges toward $\mathcal W$, proving that the ergodic metric converges to $\mathcal E^\Delta$ as the mesh discretizations are refined.
    \end{proof}

     \subsection{Ergodicity over Meshes} \label{subsec:erg_manifold}
    
 In Eq.~\eqref{eq:fk_approx}, we approximate the inner products of arbitrary functions $f$ with the basis functions $f_k$. Since the integral over $\mathcal W$ defines an inner product space, the integral can be discretized over the mesh $\mathbb{M}$ with a mass matrix $M$, also provided by \cite{Sharp:2020:LNT}.
        \begin{equation} \label{eq:fk_approx}
            \int_\mathcal{W} f(w) f_k(w) dw \approx ([f(w_i)]_{i=1}^m)^\top M \vec{f_k}
        \end{equation}
    Thus, we can approximately evaluate the integrals in Eq.~\eqref{eq:ck} and Eq.~\eqref{eq:phik} using the eigenvectors $\vec{f}_k$ of $L$.
    Our ergodic trajectory optimization (ETO) over meshable surfaces problem is:
    \begin{mini}
    {x(t),u(t)}
    {\mathcal{E}^L(x(t), \phi)}
    {\label{eq:ergodic-opt}}
    {\text{ETO}}
    \addConstraint{\dot{x}}{=f(x(t),u(t))\quad}{\forall{t}\in[0,t_f]}
    \addConstraint{r - d_{M}(x(t))}{\leq{}0\quad}{\forall{t}\in[0,t_f]}
    \addConstraint{x(t)\in{\mathcal{X}},}{u(t)\in{}\mathcal{U}\quad}{\forall{t}\in[0,t_f]}
    \end{mini}
here $f:\mathcal{X} \times \mathcal{U} \rightarrow T\mathcal{X}$ is the robot dynamics, $d_{M}:\mathcal{X}\rightarrow \mathbb{R}$ is the signed distance function of the mesh approximation of the surface that we are searching over, and $r \in \mathbb{R}^+$ is the radius of the minimum bounding sphere of the robot.

\section{Results} \label{sec:results}

In this section, we show ergodic search results over various surfaces. First, we describe the optimization problem setup (robot dynamics, constraints, algorithms to compute eigenfunctions, and optimization method). We then conduct experiments on sphere, bunny, torus, and wind turbine meshes to demonstrate our method's ability to perform both uniform and nonuniform information-driven exploration over the meshes. In this approach, as in classical ergodic trajectory optimization, the optimization is non-convex. Optimal solutions are not unique, and in practice, different initializations lead to different solution trajectories with similar objective function values \cite{miller2013trajectory,dong2024time}.

\subsection{Optimization Setup}
In our formulation, we are given a client surface model $\mathbb M\approx{}\mathcal{W}$, robot dynamics $f$, and a sensor model $s^x$ appropriate for the task. We compute the discrete operator $L$ for $\mathbb{M}$ according to \cite{Sharp:2020:LNT} and use Arnoldi methods \cite{lehoucq1998arpack} provided by SciPy to compute the eigenvalues and eigenvectors $f_k$ of $L$. These eigenvectors are used to compute the ergodic metric $\mathcal{E}^L$ Eq.~\eqref{eq:erg_met_disc}. We discretize ETO in time using forward-Euler transcription and solve the resulting nonconvex finite-dimensional optimization with an Augmented Lagrangian method \cite{birgin2014practical,lerch2023safety}.
For this paper, we choose the L-BFGS \cite{Nocedal2006} method for gradient updates. 
For the experiments presented below, we use the isotropic Gaussian sensor model given in Eq.~(\eqref{eq:sensor-model}).
\begin{equation} \label{eq:sensor-model}
    s^x(w) = \exp\left(-\frac{1}{2}\frac{(w-x)^\top (w-x)}{\sigma^2}\right)
\end{equation}

Here, $\sigma$ governs the sensor size, $x$ is the robot position, and $w$ is a query point on the client surface. The Gaussian sensor model is a common choice for coverage or inspection tasks since it provides a tractable and differentiable coverage function \cite{coffin2022multi, rao2024learning}.

Unless otherwise noted, $\mathcal{X}=\mathbb{R}^3$ (position), $\mathcal{U}=[\underline{u}_0,\bar{u}_0]\times{}[\underline{u}_1,\bar{u}_1]\times{}[\underline{u}_2,\bar{u}_2]\subset{}\mathbb{R}^3$ (speed limits along each axis), and we impose single integrator dynamics, a.k.a. velocity control:
\begin{equation}
    \dot{x}=f(x,u)=u\label{eq:singleint}
\end{equation}
Additionally, unless otherwise specified for comparison purposes, we used 100 time steps for all the trajectories with varying $dt$ values as noted below. With the exception of the unit square in $\mathbb{R}^2$, we defined our $\Lambda_k$ discount factor as:
\begin{equation} \label{eq:lambdak}
    \Lambda_k = e^{-0.1\sqrt{\lambda_k}}
\end{equation}
where $\lambda_k$ represents the eigenvalues or natural frequencies of the Laplace-Beltrami operator $L$.

\subsection{Comparison to Analytic Basis Functions}

When $\mathcal W$ is known to be a square or a sphere, we have analytic expressions to use as orthonormal bases. In these cases, we can compute the ergodic metric using the continuous Laplace-Beltrami operator, $\mathcal E^\Delta$, by using the analytic basis functions. We compare trajectories optimized for $\mathcal E^\Delta$ with trajectories optimized for our implementation of $\mathcal E^L$ (which uses the discretized operator) to demonstrate that we recover comparable performance to classical ergodic trajectory optimization. 

\begin{figure}[htb]
    \centering
    \includegraphics[width=\linewidth]{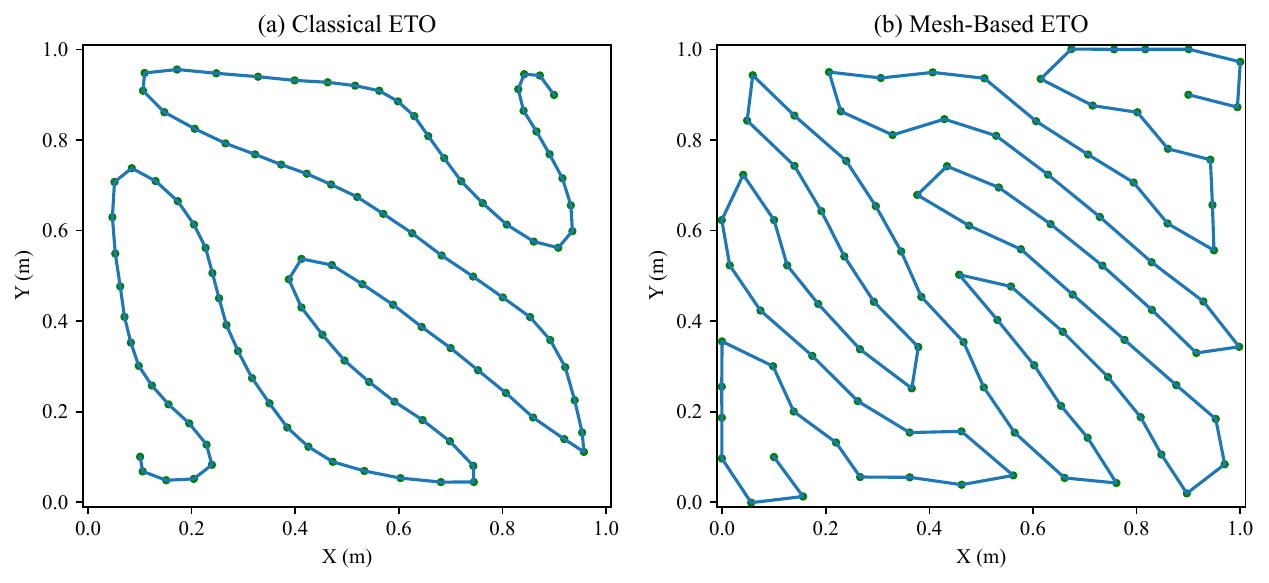}
    \vspace{-10pt}
    \caption{\textbf{Simple $\mathbb{R}^2$ Surface Search Comparison} Searching over 2D Euclidean space and comparing the (a) normal ergodic search with analytical solutions to the basis functions (ergodicity of $0.00254$) to (b) our mesh approximation and numerical methods to compute the basis functions (ergodicity of $0.00304$ with respect to the analytical basis functions)}
    \label{fig:r2}
    \vspace{-10pt}
\end{figure}

For a unit square in $\mathbb R^2$, Fig. \ref{fig:r2} shows the trajectories generated using the analytical Fourier basis functions (a) and mesh basis functions (b), both from a straight line initial trajectory connecting the initial and final position. For (b), we had $10000$ mesh vertices, sensor size $\sigma = 0.03$m, $dt = 0.1$s, the agent's velocity was constrained to less than 1 m/s, and the discount factor was instead $\Lambda_k = (1 + \sqrt{\lambda_k})^{-1}$. The trajectory's ergodic metric $\mathcal E^\Delta$ (evaluated on the analytical basis functions) in (a) is $0.00254$, while (b) achieved an ergodicity of $0.00304$. These are listed in the first block of Table \ref{tab:square-erg-metrics}.
In the table, $x_L$ refers to trajectories planned using $\mathcal E^L$, $x_\Delta$ to trajectories planned using $\mathcal E^\Delta$.
This confirms that optimizing a trajectory with $\mathcal E^L$ or $\mathcal E^\Delta$ yields similar levels of coverage for the plane.


\begin{table}[h!]
    \centering
    \caption{Comparison of ergodic metric calculated with $\Delta$ vs $L$.}
    \label{tab:square-erg-metrics}
    \begin{tabular}{S|SS}
        { } & {$\mathcal{E}^\Delta$} & {$\mathcal{E}^L$} \\ \midrule
        \text{Uniform Square } $x_\Delta$  & 0.00254  & 0.017443 \\
        \text{Uniform Square} $x_L$ & 0.00304  & 0.000493 \\
        \text{Uniform Sphere} $x_\Delta$ & 0.07180  & 0.021800 \\
        \text{Uniform Sphere} $x_L$ & 0.07620  & 0.000388
    \end{tabular}
\end{table}

\begin{figure}[htb]
    \centering
    \includegraphics[width=.9\linewidth]{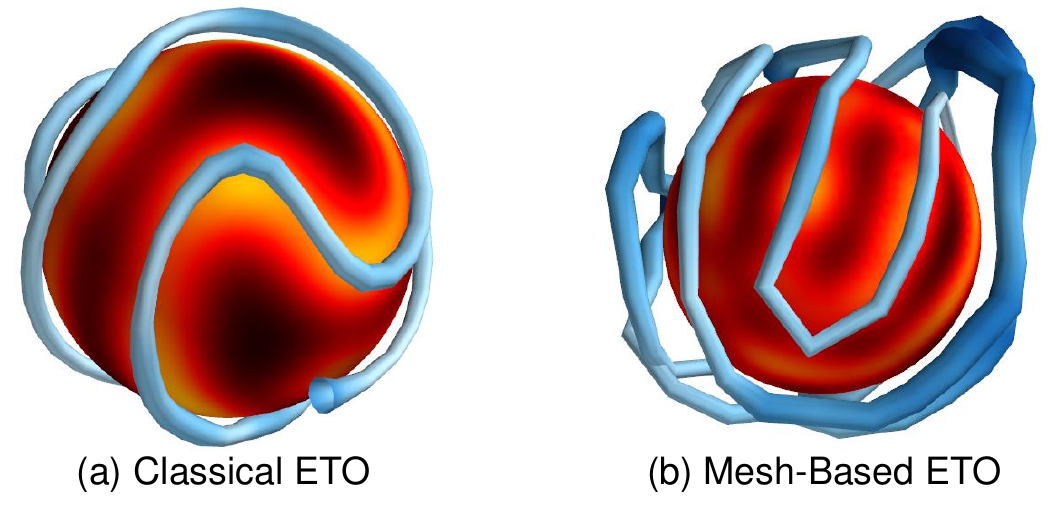}
    \vspace{-10pt}
    \caption{\textbf{Sphere Surface Search Comparison} Classical ergodic objective vs. our mesh-based ergodic objective for a sphere. The trajectory is darker/thicker to indicate lower velocity and lighter/thinner to indicate higher velocity. The information gathered is visualized as a heat map with red indicating more information. 
    The collected information is comparable with (a) the classical ETO ergodicity of $0.0718$ and (b) our proposed mesh-based ETO ergodicity of $0.0762$ with respect to the analytical basis functions.
    }
    \label{fig:sphere_comp}
\end{figure}

Next for the sphere, we compare against the trajectories obtained using the spherical harmonic basis functions \cite[\href{https://dlmf.nist.gov/14.30.E2}{(entry 14.30.E2)}]{DLMF}. For a uniform information map, Fig. \ref{fig:sphere_comp} shows the trajectories generated from the analytical basis functions (a) and the mesh basis functions (b) for a $0.5$m radius sphere from an initial trajectory that forms a ring around the center of the sphere. Here, we used 4902 mesh vertices, sensor size $\sigma = 0.1$m, $dt = 0.1$s, and the agent was constrained to stay at least $0.05$m from the mesh surface. Again, we see similar levels of coverage with ergodicities of $0.0718$ for (a) and $0.0762$ for (b) (computed using $\mathcal{E}^\Delta$) (more in second block of Table \ref{tab:square-erg-metrics}).

\subsection{Comparison to HEDAC}

\begin{figure}[htb]
    \centering
    \includegraphics[width=\columnwidth]{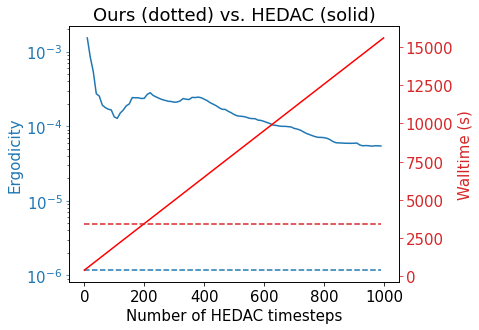}
    \caption{\textbf{HEDAC Comparison} Comparing our mesh-based ergodic search (planned for only 100 timesteps, dotted) to the HEDAC approach \cite{ivic2023multi} (with 0 to 1000 timesteps, solid) for the wind turbine mesh. Blue is ergodicity achieved; Red is the walltime used to compute the trajectories.
    }
    \vspace{-10pt}
    \label{fig:hedac_comp}
\end{figure}

\begin{figure}[htb]
    \centering
    \includegraphics[height=.8\linewidth]{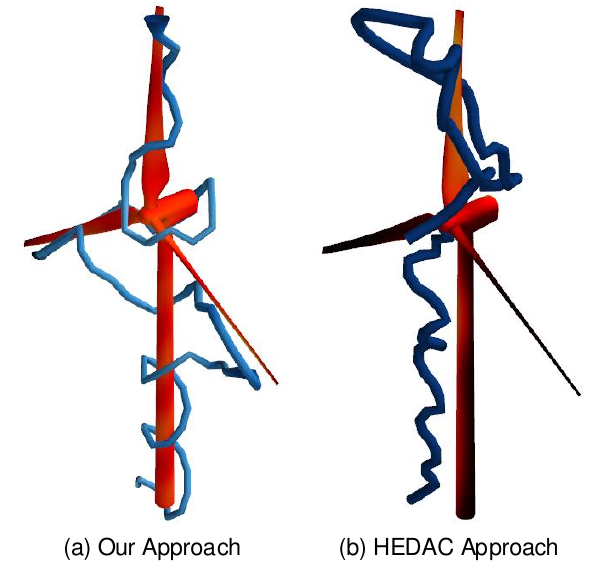}
    \caption{\textbf{Wind Turbine Ergodic Search Trajectory} (a) our mesh-based ergodic search and (b) the HEDAC approach \cite{ivic2023multi} on a wind turbine mesh for 100 timesteps. Our approach achieved an ergodicity of $1.17 \times 10^{-6}$ and HEDAC achieved an ergodicity of $5.43 \times 10^{-5}$. HEDAC leaves two of the blades unexamined.
    }
    \vspace{-10pt}
    \label{fig:windturbine}
\end{figure}
HEDAC is another non-uniform coverage trajectory generation method that can generate ergodic trajectories for complex exploration domains \cite{ivic2023multi}. We use a wind turbine mesh from \cite{ivic2023multi} to illustrate the differences from our method and highlight their relative strengths and weaknesses.

The wind turbine mesh has dimensions of $25.6 \text{m} \times 122.8 \text{m} \times 203.1 \text{m}$. The agent must stay $>5$m from the surface and move $<0.5$m/s in all directions. We used a sensor size of $\sigma = 10$m and $dt = 10$s. For our approach, we planned a trajectory of $100$ timesteps, which had an ergodicity of $1.17 \times 10^{-6}$. For HEDAC, we swept the number of timesteps from 0 to 1000, which results in decreasing ergodicity as timesteps increase (Fig. \ref{fig:hedac_comp}). Even when permitted 10x more steps than our method, it did not achieve comparable ergodicity. In Fig. \ref{fig:windturbine}, a side-by-side comparison of 100 timestep trajectories is shown. HEDAC uses 36 threads on an Intel\textregistered{} Core\texttrademark{} i9-10980XE; our approach runs on an Nvidia GeForce RTX 2080 Ti GPU.

\subsection{Additional Mesh Experiments}

\subsubsection*{\textbf{Uniform Information Distribution}}
Next, we test our method on various curved surfaces embedded in $\mathbb R^3$ to demonstrate the generality of this approach. For the following experiments, we use a 3D single-integrated robot as described in Eq.~\eqref{eq:singleint} with the same Gaussian sensor model in Eq.~\ref{eq:sensor-model}. Three meshes (sphere, torus, bunny) with uniform information distributions on their surface. The selected set of meshes tests a variety of surface topologies, details, and smoothness to demonstrate the versatility of this approach.

The sphere is the same size as above, while the torus has a radius of $0.5$m and a height of $0.285$m, and the bunny has dimensions of $1.0\text{m} \times 1.0\text{m} \times 1.0\text{m}$. All three had a sensor size of $\sigma = 0.1$m, $dt = 0.1$s were constrained to stay at least $0.05$m away from the surface, and the bunny trajectory was limited to $0.8$ m/s.

The generated trajectories are in the first row of Fig. \ref{fig:search-experiments}, given initial trajectories of a straight line from the maximum point to the minimum point. The distribution of observed information is presented as a heatmap on the surface with lighter/redder colors representing more information. The trajectory paths lines are thicker when the robot spent more time on that part of the path.

\subsubsection*{\textbf{Non-Uniform Information Distribution}}

\begin{figure}
    \centering
    \includegraphics[width=0.95\columnwidth]{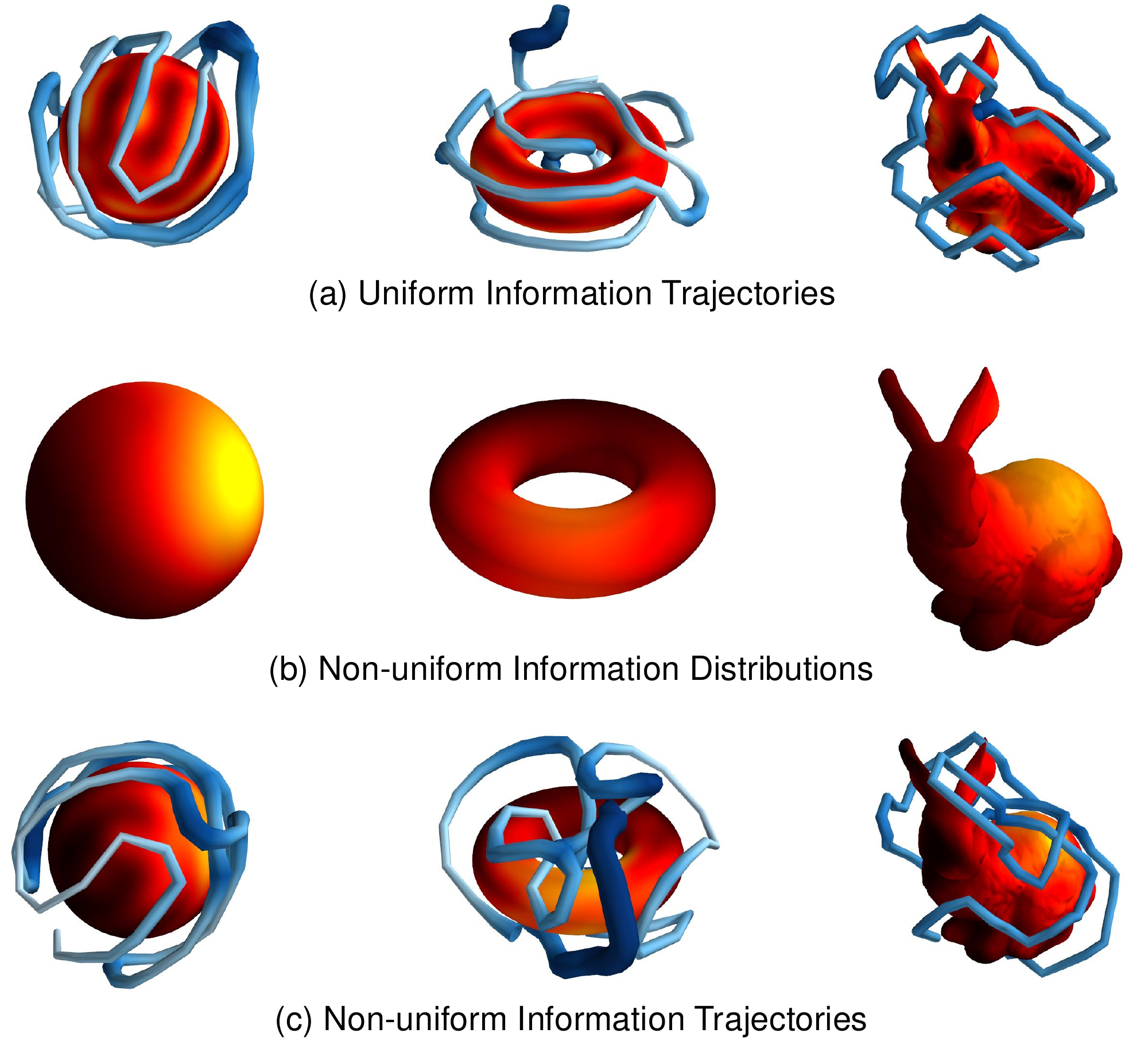}
    \caption{Search over varying information distributions on sphere, torus, and bunny meshes. (a): trajectories for a uniform information distribution. (b): single-peak non-uniform information distribution. (c): trajectories generated for the information distribution in (b). The heatmap represents gathered information in (a) and (c) and available information in (b). Whiter/thinner means faster speed, and darker/thicker means slower in trajectories.}
    \label{fig:search-experiments}
\end{figure}

Lastly, we search on sphere, torus, and bunny meshes with non-uniform information distributions. The non-uniform distributions painted on the meshes are visualized in the middle row of Fig. \ref{fig:search-experiments}, generally behaving as a single peak on the mesh (lighter/redder is more information). The trajectories found by ETO (and the information gathered) are shown at the bottom of Fig. \ref{fig:search-experiments}.

\section{Conclusion} 
\label{sec:conclusion}
In this paper, we demonstrated an ergodic search approach that can search over any surface that can be approximated as a triangle mesh, building upon previous work that was limited by a need for analytically computed basis functions for the domain. Our method shows how appropriate basis functions can be approximated using numerical methods, permitting the search of more complex domains. Where the classical ergodic search approach applies, our method achieves similar quality coverage, and in domains where the classical approach does not apply, we produce high-quality inspection trajectories.

We encounter a trade-off between the approximation accuracy of the eigenfunctions and the number of mesh vertices. As the number of mesh vertices increases, so too does the size of the discrete Laplace-Beltrami operator, thus increasing the computation time for obtaining the eigenvectors and more expensive objective function evaluations during optimization. Standard ergodic search difficulties, such as accounting for sensor noise that can disrupt the information gathering and dealing with uncertain or dynamic environments, naturally affect our method as well.

Future work will adapt this approach for execution on a physical system, considering various sensor models. Furthermore, we hope to deploy a similar search with a multi-robot system to search over objects collaboratively and more accurately model a real-world inspection procedure. 




\FloatBarrier
\bibliography{references}
\bibliographystyle{IEEEtran}

\end{document}